\documentclass[nohyperref, table]{article}

\usepackage{microtype}
\usepackage{graphicx}
\usepackage{subfigure}
\usepackage{booktabs} 

\usepackage{hyperref}

\usepackage[dynnworkshop]{icml2022} 

\usepackage{algorithmic} 
\usepackage{algorithm} 
\usepackage{graphicx}
\usepackage{amsmath}
\usepackage{amssymb}
\usepackage{booktabs} 
\usepackage{multirow} 
\usepackage{mathtools}
\usepackage{amsthm}
\usepackage{hhline} 
\usepackage{highlight} 

\usepackage[capitalize,noabbrev]{cleveref} 

\newcommand{\edits}[1]{{{\color{blue}#1}}{}} 
\newcommand{\ie}{\textit{i}.\textit{e}.}

\newcommand{\etc}{\textit{etc}}
\renewcommand{\opacity}{50}

\theoremstyle{plain}

\theoremstyle{definition}

\theoremstyle{remark}

\usepackage[textsize=tiny]{todonotes}

\icmltitlerunning{Play It Cool: Dynamic Shifting Prevents Thermal Throttling} 

\begin{document} 

\twocolumn[
\icmltitle{Play It Cool: Dynamic Shifting Prevents Thermal Throttling} 



\icmlsetsymbol{equal}{*}

\begin{icmlauthorlist}
\icmlauthor{Yang Zhou}{sch} 
\icmlauthor{Feng Liang}{sch} 
\icmlauthor{Ting-wu Chin}{sch_the_other} 
\icmlauthor{Diana Marculescu}{sch,sch_the_other} 
\end{icmlauthorlist}

\icmlaffiliation{sch}{The University of Texas at Austin} 
\icmlaffiliation{sch_the_other}{Carnegie Mellon University} 

\icmlcorrespondingauthor{Yang Zhou}{yangzhou25672@utexas.edu}

\icmlkeywords{Machine Learning, ICML}

\vskip 0.3in
]



\printAffiliationsAndNotice{}  

\begin{figure*}[t] 
  \centering \includegraphics[scale=0.55]{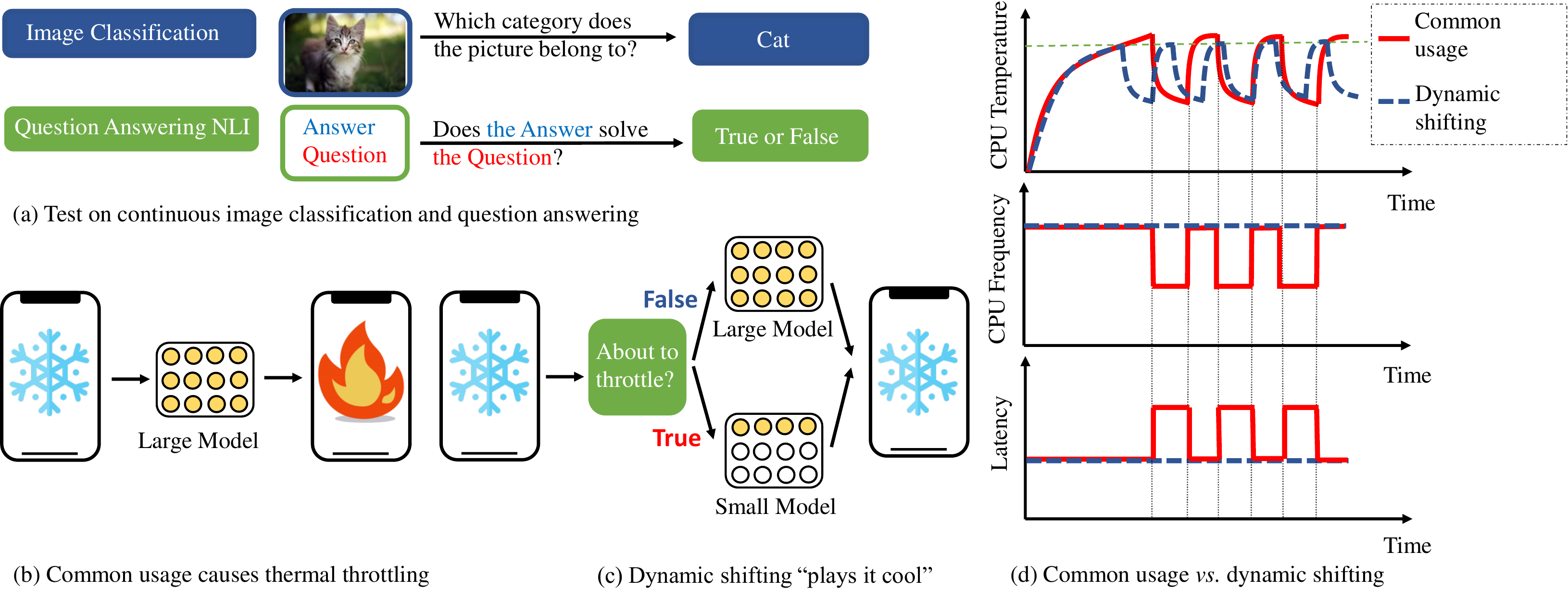}  
  \vspace{-2em}
  \caption{Comparison between common usage and our dynamic shifting. 
  (a). We adopt two widely user applications: image classification in CV and question answering natural language inference (QNLI) in NLP. 
  (b). Common usage leads to thermal throttling, resulting in an unpleasant 'hot and slow' experience.
  (c). Our method can 'play it cool' via dynamically shifting models according to the thermal profile.
  (d). Thermal throttling results in CPU frequency drop and latency increase. Our method is able to run consistently without these issues.} 
  \label{fig:main_figure}
\end{figure*}

\begin{abstract}
Machine learning (ML) has entered the mobile era where an enormous number of ML models are deployed on edge devices.
However, running common ML models on edge devices continuously may generate excessive heat from the computation, forcing the device to "slow down" to prevent overheating, a phenomenon called \emph{thermal throttling}.
This paper studies the impact of thermal throttling on mobile phones: when it occurs, the CPU clock frequency is reduced, and the model inference latency may increase dramatically. 
This unpleasant inconsistent behavior has a substantial negative effect on user experience, but it has been overlooked for a long time.
To counter thermal throttling, we propose to utilize \emph{dynamic networks} with shared weights and dynamically shift between large and small ML models seamlessly according to their thermal profile, \ie, shifting to a small model when the system is about to throttle.
With the proposed dynamic shifting, the application runs consistently without experiencing CPU clock frequency degradation and latency increase.
In addition, we also study the resulting accuracy when dynamic shifting is deployed and show that our approach provides a reasonable trade-off between model latency and model accuracy.
\end{abstract}

\section{Introduction} 
\label{sec:intro} 

ML models have become the \textit{de-facto} building blocks for virtually all computer vision (CV) and natural language processing (NLP) applications. 
Additionally, because of the expected lower inference latency and higher user data security compared to cloud servers, it has become increasingly desirable to deploy ML models on edge devices. 
However, the demanding hardware requirements of ML models has become a barrier preventing the deployment of large and powerful models on devices with limited thermal and computational resources. 
This paper focuses on an overlooked thermal issue of deploying ML models on edge devices.

When running heavy workload on edge devices, heat is continuously generated, but due to the limited thermal capacitance of the hardware platform, heat may not be adequately dissipated. 
When temperature rises above the hardware thermal threshold, thermal throttling is triggered. 
The CPU clock frequency is thereby forced to drop, thus reducing the temperature to protect hardware functionality. However, the latency increases as a result. 
To understand the impact of thermal throttling on mobile phones, we first conduct experiments on two intensive tasks: continuous image classification and continuous question answering (See Fig.~\ref{fig:main_figure}(a)).
We find that running full models continuously results in thermal throttling (Fig.~\ref{fig:main_figure}(b)).
Moreover, as shown in Fig.~\ref{fig:main_figure}(d), due to thermal throttling, the CPU clock frequency is decreased and therefore the latency of the model running on the edge device will drastically go up. 

We argue that thermal throttling poses a serious threat to mobile ML applications that are latency-critical. For example, during real-time visual rendering for video streaming or gaming, a sudden surge of processing latency per frame will have substantial negative effect on user experience. Also, modern mobile operating systems often provide special services and applications for vision impaired individuals, such as VoiceOver on iOS and TalkBack on Android. The user typically interacts with mobile phones by relying completely on speech, so the quality of these services is highly dependent on the responsiveness or the latency of the application. 

We propose to use weight sharing ~\emph{dynamic networks}~\cite{han2021dynamic,yu2018slimmable,hou2020dynabert} to prevent thermal throttling.
The proposed dynamic shifting is depicted in Fig.~\ref{fig:main_figure}(c).
Initially, the large model in the dynamic neural network suite runs. When the CPU temperature is about to exceed a set threshold, the small model in the same suite runs instead. 
Then, when the CPU temperature drops and becomes stable, the large model runs again. 
With the proposed dynamic shifting, the application will run consistently at the same latency and without experiencing CPU clock frequency degradation (see Fig.~\ref{fig:main_figure}(d)). 
Moreover, since the weights are shared, no additional memory loading is needed during model shifting. We provide details of our dynamic shifting algorithm in section~\ref{dynamic_shifting}. 

Our dynamic shifting is suitable for different types of models: convolution based slimmable networks~\cite{yu2018slimmable} and transformer based DynaBERT~\cite{hou2020dynabert}. 
Also, we demonstrate that dynamic shifting can generalize to other mobile CPU platforms by conducting the same experiments on a Raspberry Pi 4B platform. 
In addition, we study the resulting accuracy when dynamic shifting is deployed and show that our approach provides a reasonable trade-off between model latency and model accuracy. 


\vspace{-0.5em}
\section{Related Work} 


\textbf{Dynamic Thermal Management (DTM)} has
mainly focused on finding the optimal trade-off between controlling the temperature and maintaining performance through hardware-based methods such as Temperature-Tracking Dynamic Frequency Scaling ~\cite{skadron2004temperature}, Dynamic Voltage Scaling ~\cite{brooks2001dynamic, huang2000framework, donald2006techniques}, Fetch and Clock gating ~\cite{brooks2001dynamic, manne1998pipeline, sanchez1997thermal, skadron2002control}, mitigating computation ~\cite{skadron2004temperature}, hybrid DTM method ~\cite{skadron2004temperature}, \etc. These methods mainly target CPUs when running general tasks. Recently, ~\cite{benoit2020impact} have looked specifically at ML inference workloads and have proposed to use peripheral active cooling hardware on edge devices to control the temperature of the device. 
In comparison, we propose a software-only method to prevent thermal throttling by dynamically shifting between models that are part of the same dynamic network~\cite{han2021dynamic,yu2018slimmable,hou2020dynabert}. 




\begin{figure*}[t] 
\centering \includegraphics[scale = 0.53]{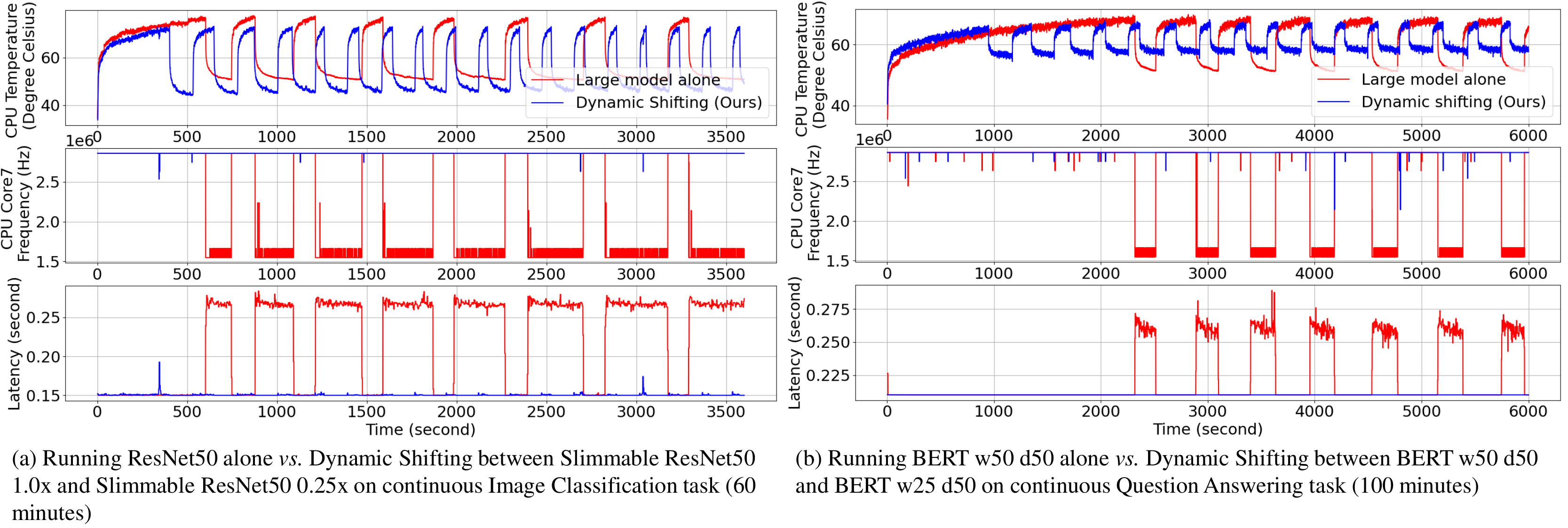} 
\vspace{-2em} 
\caption{The common usage \textit{v.s.} the proposed dynamic shifting on CV and NLP tasks on Honor V30 Pro (Best viewed in color).} 
\end{figure*} 




\setlength\textfloatsep{2pt}  

\begin{algorithm}[t] 
\small
\caption{Dynamic Shifting Algorithm} 
\label{alg:code} 
\begin{algorithmic}[1] 
\REQUIRE Dynamic Network: \textit{mModule}, Temperature Threshold: \textit{tLim}, Derivative Threshold: \textit{gLim} 
\REQUIRE setModel() sets model to small or large for \textit{mModule}; findGrad() computes the derivative; 

\STATE Initialize avgTemp, grad to 0 
\STATE mModule.setModel(LARGE) 
\WHILE{true} 
\STATE Get cpuTemp 
\STATE avgTemp $\gets \alpha \times$ avgTemp $+ (1 - \alpha) \times$ cpuTemp 
\STATE grad $\gets \beta \times \text{grad} + (1 - \beta) \times$ findGrad(avgTemp) 
\IF{mModule.inLarge() AND cpuTemp $>$ tLim} 
\STATE mModule.setModel(SMALL) 
\STATE Clear prevAvgTemp, avgTemp, grad 
\ELSIF{mModule.inSmall() AND grad $>$ gLim} 
\STATE mModule.setModel(LARGE) 
\STATE Clear prevAvgTemp, avgTemp, grad 
\ENDIF 
\ENDWHILE 
\end{algorithmic} 
\label{algorithm1} 
\end{algorithm} 

\section{Dynamic Shifting} 
\label{dynamic_shifting}

The main goal of the proposed approach is to run inference continuously within the latency constraint while not throttling. 
Once the large and small models are picked, we follow Algorithm~\ref{alg:code} to implement dynamic shifting. 
We start from the large model. When the CPU temperature exceeds a given temperature threshold, we shift to the small model. 
After shifting to the small model, the CPU temperature starts dropping. Since the small model is less computationally intensive, running the small model meets the throughput requirement. Later, when the temperature drops and becomes stable, the smart phone CPU reaches a thermal equilibrium, and the excessive heat is dissipated.  Therefore, if the temperature derivative calculated when running the small model is close enough to zero or exceeds a negative derivative threshold, we shift to the large model again for better inference accuracy. 
To compute the derivative, the CPU temperature is first smoothed out using the Exponential Moving Average (EMA). The derivative is thereafter approximated by the rate of change of the temperature readings. In practice, we use EMA again to smooth out the computed derivatives to reduce noise. 

\vspace{-1em}
\section{Experiments and Discussion} 
\paragraph{Setup, dataset, model, platform, settings.} 
We use slimmable networks~\cite{yu2018slimmable} and DynaBERT~\cite{hou2020dynabert} for CV and NLP tasks, respectively. Mobile phone experiments are conducted on the Honor V30 Pro\footnote{Phone released in 2019, CPU: Kirin 990 5G 2x2.86 GHz Cortex-A76 \& 2x2.36 GHz Cortex-A76 \& 4x1.95 GHz Cortex-A55, CPU frequency is reported on core 7.}. 
To run model inferences on the phone, we encapsulate PyTorch models into TorchScript files and build testing programs upon the published Android Studio code\footnote{https://github.com/pytorch/android-demo-app}. To test the generability of our dynamic shifting, we also conduct experiments on Raspberry Pi 4B 4GB using onnxruntime. 

We find that the PyTorch torchscript can only trace a static computation graph of the model and is not suitable for shared weight dynamic networks. We have to trace the large and the small models in two separate files. Once encapsulated in the torchscript, the large and the small model no longer share weights, and because of this, extra memory overhead is noted when shifting between models. We report the time taken to load in models during shifting in Appendix~\ref{shift_and_log}. 
Moreover, we want to emphasize that if true weight sharing\footnote{We found that now there is hardly a mobile framework that supports shared weights dynamic networks. Currently, to implement a true weight sharing dynamic network on mobile devices, one must directly write a model-specific C++ or Java program.} is used, the extra memory overhead will not occur.

\begin{table}[t]
\small
\caption{Summary of the latency and accuracy of running Dynamic Shifting on Slimmable ResNet50 and DynaBERT on Honor V30 Pro (DS refers to our method Dynamic Shifting)} 
\vspace{1em}
\label{tab:acc_lat_comp} 
\begin{tabular}{l|ll} 
\hline 
& Latency (s)        & Accuracy       \\ 
\hline
\begin{tabular}[c]{@{}l@{}}
ResNet50 1.0x \\ 
(alone) 
\end{tabular}
& 0.205           & 0.761          \\ 
\begin{tabular}[c]{@{}l@{}}
Slimmable ResNet50 \\
(DS) 
\end{tabular} 
& 0.150 (-26.8\%) & 0.695 (-0.066) \\ 
\hline 
\begin{tabular}[c]{@{}l@{}} 
BERT d0.5 w0.5 \\ 
(alone + 1.4x latency) 
\end{tabular} & 0.217           & 0.900          \\ 
\begin{tabular}[c]{@{}l@{}}
DynaBERT \\ 
(DS + 1.4x latency) 
\end{tabular}          
& 0.211 (-2.88\%) & 0.893 (-0.008) \\ 
\hline
\end{tabular}
\end{table} 

\begin{figure*}[t] 
\label{RaspberryPi4} 
\centering \includegraphics[scale=0.52]{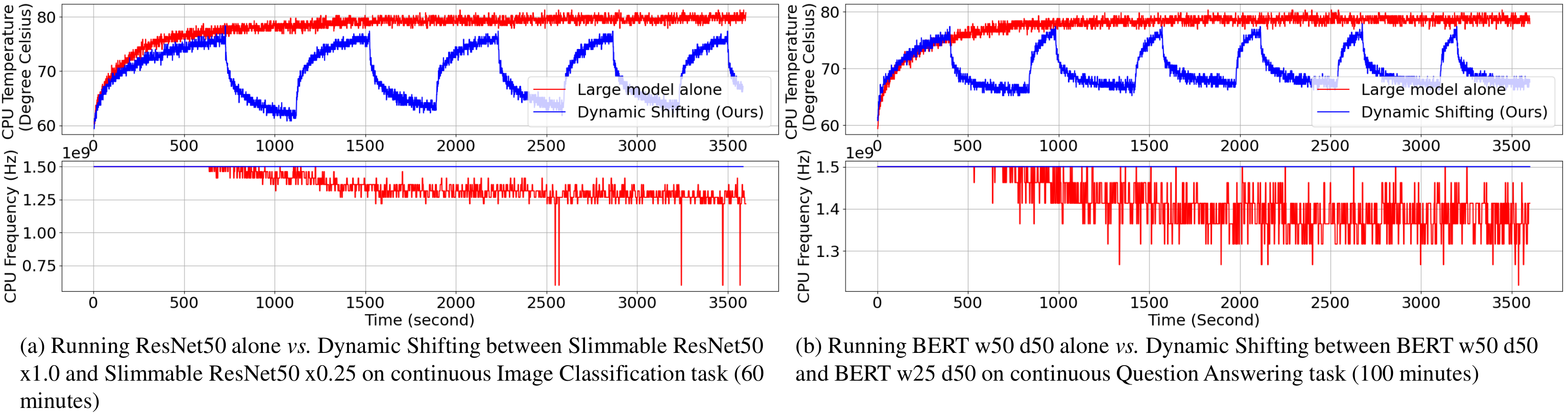} 
\vspace{-2em}
\caption{The common usage \textit{v.s.} the proposed dynamic shifting on CV and NLP tasks on Raspberry Pi 4B (Best viewed in color).} 
\end{figure*} 

\subsection{Mobile Phone Experiments} 
To make the comparison to running the large model alone easier, we fix the inference latency of the small model to have the same inference latency as the large model by injecting idle time after every small model inference. We find that the mobile phone throttling point is determined by multiple factors not only by the temperature. 
For this work, we simplify the problem and only consider using CPU temperature when designing the shifting threshold. 
Time taken for logging the data is measured and reported in Appendix~\ref{shift_and_log}. 
We document the experiment details in Appendix~\ref{experiment_details}. 

\subsubsection{Slimmable ResNet50 Experiments} 
Running ResNet50 alone throttles the device roughly at CPU temperature 77$ ^\circ C $ (see the red line in Figure 2 (a)). Starting from this, we set the temperature threshold to 73$ ^\circ C $. The derivative threshold after EMA is set to -0.07.  We shift between the largest and the smallest model in slimmable ResNet50.
From Figure 2 (a), dynamic shifting is shown to prevent thermal throttling. The CPU temperature and Latency remain stable throughout the entire hour of the experiment. 

We also report the accuracy and latency comparison in Table~\ref{tab:acc_lat_comp}. The latency is calculated by averaging all inference latency during the experiments. The accuracy is estimated by the ratio of large and small model. Although it may sacrifice some accuracy, the proposed dynamic shifting has a faster inference speed. Most importantly, our dynamic shifting approach enjoys a consistent inference. We also provide ablation studies on the choice of different combinations of temperature and derivative thresholds in Appendix~\ref{threshold_and_temperature}. 

\subsubsection{DynaBERT Experiments} 
We find that BERT models~\cite{devlin2018bert} in general are too computationally intensive for mobile phones. (see Appendix ~\ref{experiment_details}). 
To test our approach, we choose to dynamically shift between BERT d0.5 w0.5 and BERT d0.25 w0.5, the two smallest models in the DynaBERT suite. 
To make the smallest model capable to reaching a suitable operating temperature, we increase latency to 1.4X the original BERT d0.5 w0.5 latency. 
With 1.4X latency, BERT d0.5 w0.5 throttles at 70$ ^\circ C $. Accordingly, the temperature threshold is set to 65$ ^\circ C $. The derivative threshold is set to -0.008. 
We compare dynamic shifting with running BERT d0.5 w0.5 alone at the same latency is in Figure 2 (b). 
The CPU frequency and the CPU latency also remain stable for the whole 100 minutes in the experiment. The latency and accuracy trade-off is also shown in Table~\ref{tab:acc_lat_comp}.
Dynamic Shifting, in general, cannot prevent BERT models from thermal throttling because of the model's enormous computational intensity. However, under some limitations, dynamic shifting can still be helpful when deploying BERT models on mobile phones.

\subsection{Raspberry Pi 4B Experiments} 
We conduct the same experiments on Raspberry Pi 4B, aiming to show that dynamic shifting can generalize well to other mobile CPU platforms. We switch between ResNet50 1.0x and ResNet50 0.25x and between BERT d0.5 w1.0 and BERT d0.5 w0.25. Experiment setup details are in Appendix~\ref{experiment_details}. Unlike mobile phones, when Raspberry Pi 4B throttles, it maintains the CPU temperature at the throttling temperature by slightly lowering the frequency as shown in Figure 3 (a). Therefore, after throttling, the latency for ResNet50 rises only by 4.5\% and 5.7\% for BERT w1.0 d0.5. 
We argue that Raspberry Pi 4B's thermal throttling strategy is not applicable to mobile phones since the high temperature is not suitable to wearable devices. 
Nevertheless, we present the temperature and the frequency results in Figure~\ref{RaspberryPi4} and show that dynamic shifting can stop Raspberry Pi 4B from throttling. 

\vspace{-0.5em}
\section{Conclusion And Future Work} 
In this work, we show that thermal throttling is a serious problem in deploying common neural networks on mobile phone. To counter thermal throttling, we propose to use dynamic shifting between models in dynamic networks with shared weights based on the device CPU temperature. Through the experiments, we demonstrate the effectiveness of dynamic shifting in Computer Vision and Natural Language Processing tasks under some conditions. 

In this work, we only consider shifting between large and small models. Further attempts can incorporate models with sizes in between into the shifting iteration. Also, future work that takes multiple factors into account when designing threshold for shifting between models can potentially perform more efficiently than ours. Besides, dynamic shifting by its own is not sufficient for models that are compute intensive such as BERT models, because even running only the smallest model would not make the phone operate under suitable temperature. We leave the problem for future work. 

\vspace{-1em}
\section{Acknowledge} 
The research is supported in part by NSF CCF Grant No. 2107085 and NSF CSR Grand No. 1815780. 


\bibliography{egbib}

\begin{thebibliography}{13}
\providecommand{\natexlab}[1]{#1}
\providecommand{\url}[1]{\texttt{#1}}
\expandafter\ifx\csname urlstyle\endcsname\relax
  \providecommand{\doi}[1]{doi: #1}\else
  \providecommand{\doi}{doi: \begingroup \urlstyle{rm}\Url}\fi

\bibitem[Benoit-Cattin et~al.(2020)Benoit-Cattin, Velasco-Montero, and
  Fern{\'a}ndez-Berni]{benoit2020impact}
Benoit-Cattin, T., Velasco-Montero, D., and Fern{\'a}ndez-Berni, J.
\newblock Impact of thermal throttling on long-term visual inference in a
  cpu-based edge device.
\newblock \emph{Electronics}, 9\penalty0 (12):\penalty0 2106, 2020.

\bibitem[Brooks \& Martonosi(2001)Brooks and Martonosi]{brooks2001dynamic}
Brooks, D. and Martonosi, M.
\newblock Dynamic thermal management for high-performance microprocessors.
\newblock In \emph{Proceedings HPCA Seventh International Symposium on
  High-Performance Computer Architecture}, pp.\  171--182. IEEE, 2001.

\bibitem[Devlin et~al.(2018)Devlin, Chang, Lee, and Toutanova]{devlin2018bert}
Devlin, J., Chang, M.-W., Lee, K., and Toutanova, K.
\newblock Bert: Pre-training of deep bidirectional transformers for language
  understanding.
\newblock \emph{arXiv preprint arXiv:1810.04805}, 2018.

\bibitem[Donald \& Martonosi(2006)Donald and Martonosi]{donald2006techniques}
Donald, J. and Martonosi, M.
\newblock Techniques for multicore thermal management: Classification and new
  exploration.
\newblock \emph{ACM SIGARCH Computer Architecture News}, 34\penalty0
  (2):\penalty0 78--88, 2006.

\bibitem[Han et~al.(2021)Han, Huang, Song, Yang, Wang, and
  Wang]{han2021dynamic}
Han, Y., Huang, G., Song, S., Yang, L., Wang, H., and Wang, Y.
\newblock Dynamic neural networks: A survey.
\newblock \emph{IEEE Transactions on Pattern Analysis and Machine
  Intelligence}, 2021.

\bibitem[Hou et~al.(2020)Hou, Huang, Shang, Jiang, Chen, and
  Liu]{hou2020dynabert}
Hou, L., Huang, Z., Shang, L., Jiang, X., Chen, X., and Liu, Q.
\newblock Dynabert: Dynamic bert with adaptive width and depth.
\newblock \emph{Advances in Neural Information Processing Systems},
  33:\penalty0 9782--9793, 2020.

\bibitem[Huang et~al.(2000)Huang, Renau, Yoo, and
  Torrellas]{huang2000framework}
Huang, M., Renau, J., Yoo, S.-M., and Torrellas, J.
\newblock A framework for dynamic energy efficiency and temperature management.
\newblock In \emph{Proceedings of the 33rd annual ACM/IEEE international
  symposium on Microarchitecture}, pp.\  202--213, 2000.

\bibitem[Manne et~al.(1998)Manne, Klauser, and Grunwald]{manne1998pipeline}
Manne, S., Klauser, A., and Grunwald, D.
\newblock Pipeline gating: Speculation control for energy reduction.
\newblock In \emph{Proceedings. 25th Annual International Symposium on Computer
  Architecture (Cat. No. 98CB36235)}, pp.\  132--141. IEEE, 1998.

\bibitem[Sanchez et~al.(1997)Sanchez, Kuttanna, Olson, Alexander, Gerosa,
  Philip, and Alvarez]{sanchez1997thermal}
Sanchez, H., Kuttanna, B., Olson, T., Alexander, M., Gerosa, G., Philip, R.,
  and Alvarez, J.
\newblock Thermal management system for high performance powerpc/sup
  tm/microprocessors.
\newblock In \emph{Proceedings IEEE COMPCON 97. Digest of Papers}, pp.\
  325--330. IEEE, 1997.

\bibitem[Skadron et~al.(2002)Skadron, Abdelzaher, and Stan]{skadron2002control}
Skadron, K., Abdelzaher, T., and Stan, M.~R.
\newblock Control-theoretic techniques and thermal-rc modeling for accurate and
  localized dynamic thermal management.
\newblock In \emph{Proceedings Eighth International Symposium on High
  Performance Computer Architecture}, pp.\  17--28. IEEE, 2002.

\bibitem[Skadron et~al.(2004)Skadron, Stan, Sankaranarayanan, Huang, Velusamy,
  and Tarjan]{skadron2004temperature}
Skadron, K., Stan, M.~R., Sankaranarayanan, K., Huang, W., Velusamy, S., and
  Tarjan, D.
\newblock Temperature-aware microarchitecture: Modeling and implementation.
\newblock \emph{ACM Transactions on Architecture and Code Optimization (TACO)},
  1\penalty0 (1):\penalty0 94--125, 2004.

\bibitem[Wang et~al.(2018)Wang, Singh, Michael, Hill, Levy, and
  Bowman]{wang2018glue}
Wang, A., Singh, A., Michael, J., Hill, F., Levy, O., and Bowman, S.~R.
\newblock Glue: A multi-task benchmark and analysis platform for natural
  language understanding.
\newblock \emph{arXiv preprint arXiv:1804.07461}, 2018.

\bibitem[Yu et~al.(2018)Yu, Yang, Xu, Yang, and Huang]{yu2018slimmable}
Yu, J., Yang, L., Xu, N., Yang, J., and Huang, T.
\newblock Slimmable neural networks.
\newblock \emph{arXiv preprint arXiv:1812.08928}, 2018.

\end{thebibliography}
\bibliographystyle{icml2022}


\newpage
\appendix
\onecolumn


\section{Time taken to shift between models and logging} 
\label{shift_and_log} 
Since TorchScript cannot trace dynamic networks, weights are not shared between the large and the small networks. We report the extra load latency from both the phone and Raspberry Pi 4B in Table~\ref{latency_table}. The phone load models from PyTorch Torchscripts, while the Raspberry Pi 4B uses Onnxruntime files, which internally uses PyTorch Torchscript to encapsulate PyTorch models. We give both the mean and the standard deviation of the measurements. In each entry, the left value stands for the latency overhead for loading the large model, while the right value stands for the latency overhead for loading the small model. 

Please note that for DynaBERT, phone uses Bert w0.5 d0.5 as the large model and Bert w0.25 d0.5 as the small model, while the Raspberry Pi 4B uses Bert w1.0 d0.5 as the large model and Bert w0.25 d0.5 as the small model. Both phone and Raspberry Pi 4B uses Slimmable ResNet50 1.0x as the large model and Slimmable ResNet50 0.25x as the small model. 

\begin{table}[h] 
\centering 
\caption{Extra latency from shifting between models (large/small)} 
\label{latency_table} 
\vspace{1em}
\begin{tabular}{c|cc} 
\hline 
\multirow{2}{*}{} & \multicolumn{2}{c}{Shifting Latency (s)} \\
& Phone & Raspberry Pi \\ 
\hline 
\begin{tabular}[c]{@{}l@{}}  
Slimmable \\ ResNet50 
\end{tabular} & 1.000 $\pm$ 0.252/0.997 $\pm$ 0.321 & 0.887 $\pm$ 0.070/0.143 $\pm$ 0.006 \\ 
DynaBERT & 0.885 $\pm$ 0.066/0.826 $\pm$ 0.016 & 1.527 $\pm$ 0.423/0.810 $\pm$ 0.218 \\ 
\hline 
\end{tabular}
\end{table} 

While running the models, we log CPU temperature, CPU frequency, and inference latency after every inference. We report the extra gap time from logging (without logging the memory usage) from both the phone and from Raspberry Pi 4B in Table~\ref{latency_table}. We provide both the mean and the standard deviation of the measurements. 

\begin{table}[h] 
\centering 
\caption{Average time taken for logging} 
\vspace{1em}
\label{logging_latency_table} 
\begin{tabular}{c|c} 
\hline 
& Logging Latency (s) \\
\hline
Phone & 0.023 $\pm$ 0.004 \\ 
Raspberry Pi 4B & 0.080 $\pm$ 0.014 \\ 
\hline 
\end{tabular} 
\end{table} 

\section{Experiment Details} 
\label{experiment_details} 
\subsection{Mobile Phone Slimmable ResNet50} 
We select a random 224-by-224 image and run inferences of the static image continuously with no gap between consecutive inferences. 
The CPU temperature is read from the built-in sensor of the phone. The CPU temperature is read after every inference, together with CPU frequency and the model inference latency. We use the official PyTorch implementation of Slimmable ResNet50\footnote{\texttt{https://github.com/JiahuiYu/slimmable\_networks}}. The EMA coefficients $\alpha$ and $\beta$ in Algorithm~\ref{algorithm1} are chosen to be 0.995 and 0.99 respectively. When running all experiments on the phone, we keep the atmosphere in room temperature around 22$ ^\circ C $, and we make sure the phone is unplugged and has battery level over 70\%. \edits{Besides, we kill all other tasks except the experiment app while running the experiments.} 
\subsection{Mobile Phone DynaBERT} 
We randomly select one question answer pair from the QNLI dataset from GLUE~\cite{wang2018glue} as input for every iteration of inferences. The total sum of input tokens of the question and answer pair is 128, the same as all other pairs in the QNLI dataset. Similar to the Slimmable ResNet50 experiments, we let the phone run inference continuously with no gap in between inputs. We use the official PyTorch implementation of DynaBERT\footnote{\texttt{https://github.com/huawei-noah/Pretrained-Language-Model/tree/master/DynaBERT}}. The EMA coefficients we used during experiments are the same as in Slimmable ResNet50 experiments. 

As we report in the paper, BERT models in general are computationally intensive to mobile phone CPUs. The Honor V30 Pro CPU temperature rises  to 80$ ^\circ C $ in less than 32 seconds and experiences thermal throttling in less than 6 minutes when running the full BERT d1.0, w1.0 (depth 1.0x, width 1.0x). A similar trend is seen for other BERT models with either full depth or full width. Therefore, we choose to only use BERT models with half the width. 
\begin{figure*}[th] 
\label{mem_us_plot} 
\centering 
\includegraphics[scale=0.49]{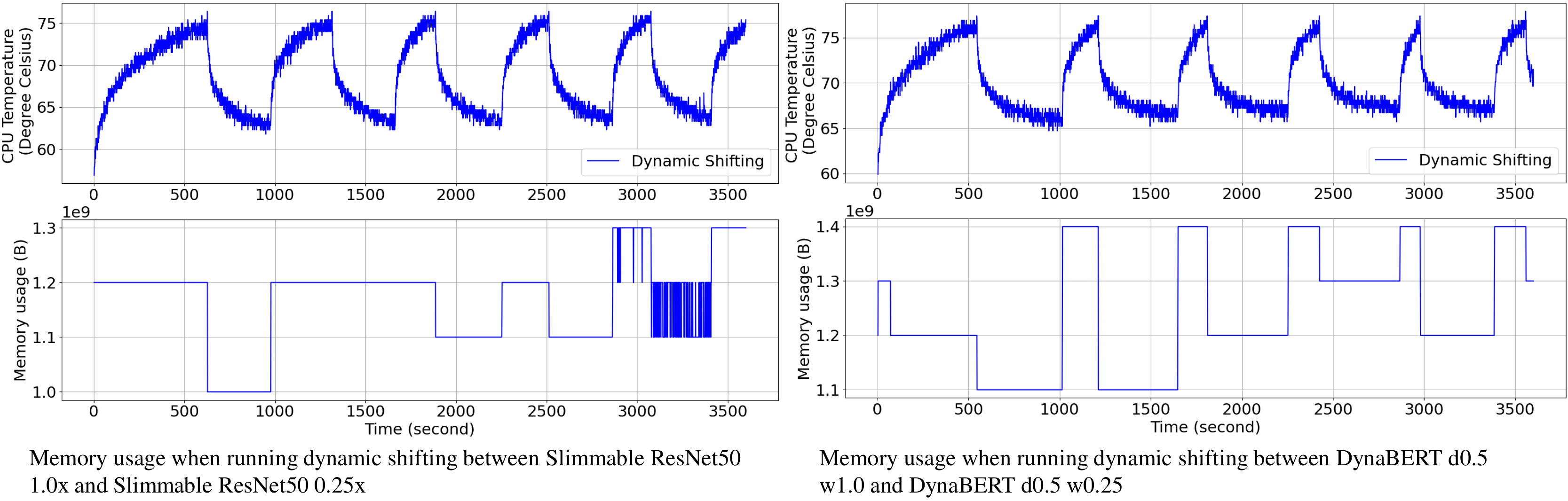} 
\vspace{-1em} 
\caption{The memory overhead in Bytes throughout 3600 seconds of Dynamic Shifting on Raspberry Pi 4B (temperature plots might look slight different compared to plots in the body)} 
\end{figure*} 
\subsection{Raspberry Pi Slimmable ResNet50} 
We measure the CPU temperature by reading the built-in sensor inside Raspberry Pi 4B. Raspberry Pi 4B has a fixed thermal throttling point which can be modified to any value below 85$ ^\circ C $. To match the mobile phone setting, we set the thermal throttling point to 80$ ^\circ C $. We observe from the experiments that Raspberry Pi 4B operating system starts to throttle the frequency at around 78$ ^\circ C $. Therefore, we set the temperature threshold to 77$ ^\circ C $. For running dynamic shifting on Slimmable ResNet50, we set the derivative threshold at -0.02. We use the same EMA coefficients as in Mobile Phone experiments, namely, $\alpha$ and $\beta$ to be 0.995 and 0.99 respectively. 
\subsection{Raspberry Pi DynaBERT} 
As in Slimmable ResNet50 experiments, we set the temperature threshold to be 77$ ^\circ C $. We set the derivative threshold to be -0.012, higher than the derivative threshold in Slimmable ResNet50 to give more time to run the small model to cool down the CPU. 

In the case of mobile phones, we find that using BERT d0.5 w1.0 model can throttle the CPU very quickly. In contrast, we find that it is not the case in Raspberry Pi 4B. Therefore, we shift between BERT d0.5 w1.0 and BERT d0.5 w0.25. 
\section{Ablation Study} 
\label{threshold_and_temperature} 

\begin{table}[ht]
    \centering
    \caption{Accuracy comparison between different combinations of temperature and derivative thresholds (Best viewed in color)} 
    \vspace{1em} 
    \label{tab:mytab}
            
    \begin{tabular}{cc|cccc} 
    \hline 
    \multirow{2}{*}{} & \multirow{2}{*}{} & \multicolumn{4}{c}{Temperature Threshold $ ^\circ C $} \\ 
    && 75 & 73 & 70 & 65\\ 
    \hline 
    & -0.005 & \gradient{0.693} & \gradient{0.684} & \gradient{0.675} & \gradient{0.668}\\ 
    Derivative & \gradient{-0.01}  & \gradient{0.700} & \gradient{0.688} & \gradient{0.678} & \gradient{0.667}\\ 
    Threshold & -0.07 & \gradient{0.710} & \gradient{0.701} & \gradient{0.684}      & \gradient{0.671} \\ 
    & -0.10 & \gradient{0.716} & \gradient{0.698} & \gradient{0.692}      & \gradient{0.677} \\ 
\hline 
\end{tabular} 
\end{table} 

The temperature threshold and the derivative threshold are two hyperparameters in dynamic shifting. Different choices of the temperature and derivative thresholds are explored. We run dynamic shifting using Slimmable ResNet50 1.0x and Slimmable ResNet50 0.25x for 30 minutes. To compare hyperparameters, we compute the average theoretical Top-1 accuracy of the first two stable shifting iterations. The models' accuracy is based on the Slimmable Networks paper (0.768 for model 1.0x and 0.638 for model 0.25x). The results are shown in Table 4. 

The general trend in the Table 4 is that the accuracy is the highest in the lower left corner and the lowest in the upper right. Towards the lower left, the temperature threshold is the highest and the derivative threshold is the lowest. The large model can potentially run longer, and the small model will reach the derivative threshold sooner. We choose 73$ ^\circ C $ rather than 75$ ^\circ C $ for the experiment in Figure 2 (a) to be more conservative towards thermal throttling. Also, we find that using 75$ ^\circ C $ makes the time running large model uneven between iterations, which can explain why in Table 1 75$ ^\circ C $, -0.10 has average accuracy slightly lower than the accuracy of 73$ ^\circ C $, -0.07. 

\section{Memory Overhead} 
\label{memory_usage}

We include the memory usage plots from the Raspberry Pi 4B during the dynamic shifting in Figure 4. Please note that since logging memory usage increases the logging latency, experiments are run separately. The memory usage is not logged in the experiments presented in the body. The temperature graphs might look slightly different here than in Figure~\ref{RaspberryPi4} as a result.


\end{document}